\documentclass[preprint]{elsarticle}

\usepackage{amssymb,amsmath,amsthm}
\usepackage{graphicx,psfrag,epsf}
\usepackage{enumerate}
\usepackage{natbib}
\usepackage{cancel}
\usepackage[scr]{rsfso}
\usepackage[final]{pdfpages}
\usepackage{subcaption}
\usepackage{booktabs}
\usepackage{makecell}
\usepackage{multirow}
\usepackage{natbib}
\usepackage{url} 
\usepackage[margin=1in]{geometry}
\usepackage{listings}
\usepackage{xcolor}

\journal{Computational Statistics \& Data Analysis}

\begin{document}

\begin{frontmatter}



\title{mSHAP: SHAP Values for Two-Part Models}


\author[inst1]{Spencer Matthews\corref{cor1}}

\author[inst1]{Brian Hartman}

\cortext[cor1]{Corresponding Author, email: spencm1@uci.edu}

\affiliation[inst1]{organization={Department of Statistics, Brigham Young University},
            city={Provo},
            postcode={84602}, 
            state={UT},
            country={Utah County}}

\begin{abstract}
Two-part models are important to and used throughout insurance and actuarial science. Since insurance is required for registering a car, obtaining a mortgage, and participating in certain businesses, it is especially important that the models which price insurance policies are fair and non-discriminatory. Black box models can make it very difficult to know which covariates are influencing the results. SHAP values enable interpretation of various black box models, but little progress has been made in two-part models.
In this paper, we propose mSHAP (or multiplicative SHAP), a method for computing SHAP values of two-part models  using the SHAP values of the individual models.
This method will allow for the predictions of two-part models to be explained at an individual observation level.
After developing mSHAP, we perform an in-depth simulation study.
Although the kernelSHAP algorithm is also capable of computing approximate SHAP values for a two-part model, a comparison with our method demonstrates that mSHAP is exponentially faster.
Ultimately, we apply mSHAP to a two-part ratemaking model for personal auto property damage insurance coverage. Additionally, an R package (mshap) is available to easily implement the method in a wide variety of applications.
\end{abstract}




\end{frontmatter}


\section{Introduction}
\label{sec:intro}
One of the most popular families of machine learning models are tree-based algorithms, which use the concept of many decision trees working together to create more generalized predictions \citep{lundberg2020local}. Current implementations include random forests, gradient boosted forests, and others.
These models are very good at learning relationships and have proven highly accurate in diverse areas.
Already, many aspects of life are affected by these algorithms as they have been implemented in business, technology, and more.

As these methods become more abundant, it is crucial that explanations of model output are easily available.
The exact definition of ``explanation'' is a subject of debate, and \citet{lipton2018mythos} argues that the word is often used in a very unscientific way due to the confusion over its meaning.
In this paper, we will regard an explainable system as what \citet{doran2017does} refer to as a comprehensible system, or one that ``allow[s] the user to relate properties of the inputs to their output.''

Explainable models are important not only because some industries require them, but also because understanding the why behind the output is essential to avoiding possible pitfalls. Understanding the reasoning behind model output allows for recognition of model bias, and increased security against harmful models being put into production. When implemented well, machine learning models can be more accurate than traditional models. However, more accurate model families can be less explainable, simply because of the nature of these algorithms. 
Generally, as predictive performance increases so too does model complexity, decreasing the ability to understand the effects of inputs on the output \citep{gunning2017explainable}.%

In this paper, we propose a methodology for explaining two-part models, which expands on the already prevalent TreeSHAP algorithm \citep{lundberg2020local}.
This methodology, called mSHAP, will allow the output of two models to be multiplied together while maintaining explainability of the resulting prediction.

In order to understand the implications and advantages of mSHAP, we will first revisit the existing SHAP-based methods and discuss where issues arise in the context of two-part models.
We will then propose a context in which SHAP values for two existing models can be combined to explain a two-part model.
Although this framework is robust, it does leave a part (which we call $\alpha$) of the ultimate prediction that must be distributed back into the contributions of the variables.
To this end, we run a simulation across different methods of distributing $\alpha$ and score the methods in comparison to kernelSHAP, an existing method for estimating explanations of any type of model.
Included in this simulation is a comparison of mSHAP to kernelSHAP, which highlights the advantages of mSHAP.
Having scored these methods, we select the best one, and apply the process of mSHAP on an auto insurance dataset.

\section{Motivation}
\label{sec:motivation}

The initial idea for this methodology came due to the problem of machine learning in auto insurance ratemaking (or pricing). Actuaries are tasked with taking historical data and using it to set current rates for insured consumers. Given the sensitive nature of the data and the potential impact it has to bias rates for different types of people, there are strict regulations on the models. The outputs of these models must be explainable so that regulators in the insurance industry can be sure that the rates are not unfairly discriminatory.

Many actuaries use a two-part model to set rates, where the first part predicts how many claims a policyholder will have (the claim frequency) and the second part predicts the average cost of an individual claim \citep{heras2018application, frees2010household}.
Multiplying the two outputs of these models predicts the total cost of a given policyholder. 

Two-part models are more difficult to explain than standard models, but the complexity increases when the two models themselves are not traditional generalized linear models.
Given this difficulty and the strict requirements of the regulators, machine learning models are not often used in actuarial ratemaking.
Despite the lack of current industry use, machine learning models such as tree-based algorithms could improve the accuracy of ratemaking models \citep{akinyemi2020use}.
Since the data that actuaries work with is typically tabular, tree-based algorithms are a good fit for predicting on the data.
In recent years there have been many advances in explaining tree based machine learning algorithms, which could lead to greater adaptation in the field. 
One of the most important is the SHAP value.

\subsection{SHAP Values and Current Implementations}
\label{subsec:motivation1}

SHAP values originate in the field of economics, where they are used to explain player contributions in cooperative game theory.
Proposed by \citet{shapley1953value}, they predict what each player brings to a game. 
This idea was ported into the world of machine learning by \citet{NIPS2017_7062}.
The basic algorithm calculates the contribution of a variable to the prediction for every possible ordering of variables, then averages those contributions.
This becomes computationally impractical very quickly, but \citeauthor{NIPS2017_7062} created a modified algorithm that approximates these SHAP values.

A couple years later \citet{lundberg2020local} published a new paper detailing a method called TreeSHAP.
This method is a rapid way of computing exact SHAP values for any tree-based machine learning model.
The fixed structure of trees in a tree-based model allows shortcuts to be taken in the computation of SHAP values, which greatly speeds up the process. With this improvement, it becomes feasible to explain millions of predictions from tree based machine learning algorithms.
These local explanations can then be combined to create an understanding of the whole model.

\subsection{Properties of SHAP Values}
\label{subsec:math1}

There are three essential properties of SHAP values: local accuracy/efficiency, consistency/monotonicity, and missingness \citep{NIPS2017_7062}.
These three properties are satisfied by the equation used to calculate SHAP values, as implemented by \citeauthor{NIPS2017_7062}.
While we focus on the local accuracy property for the rest of this section, we note that since mSHAP is built on top of treeSHAP, it automatically incorporates the consistency/monotonicity and missingness properties.

\subsubsection{Local Accuracy in Implementation}

The most important of the above mentioned properties in the context of mSHAP is the property of local accuracy/efficiency.
In the context of machine learning, this property says that the contributions of the variables should add up to the difference between the prediction and the average prediction of the model. The average prediction can be thought of as the model bias term, which is what the model will predict, on average, across all inputs (assuming representative training data).
Note that in the TreeSHAP algorithm the average prediction of the model is computed as the mean of all predictions for the training data set.
The SHAP values are then computed to explain deviance from the average prediction.

Thus, given an arbitrary model $Y$ with prediction $\hat{y}$ based on two predictors, $x_1$ and $x_2$, we can represent the mean prediction with $\mu_Y$ and the SHAP values for the two covariates as $s_{x_1}$ and $s_{x_2}$.
Based on the property of local accuracy, we know that $\hat{y} = \mu_Y + s_{x_1} + s_{x_2}$. 

This principle applies to models with any number of predictors and is very desireable in explainable machine learning \citep{arrieta2020explainable}.

\subsubsection{The Problem of Local Accuracy}

Since it is so important that the SHAP values add up to the model output, any attempt at explaining two-part model output from the SHAP values of the individual parts must maintain this property.
However, multiplying the output of two models blends the contributions from different variables, making it unclear what contributions should be given to what variables.
The idea of combining models, and using the SHAP values of the individual models to get the SHAP values for the combined model has been implemented before.
In a related github issue, Scott Lundberg assures that averaging model output is compatible with averaging SHAP values, as long as the SHAP values (and model output) are in their untransformed state \citep{shapIss}.
Even though averaging SHAP values for each variable works when averaging model outputs, the same principle does not apply when multiplying model outputs.

When considered, this is apparent.
In the most simple of cases, we see that if we have two models that both predict some outcome based on two covariates $x_1$ and $x_2$, we can average their results and likely get a better prediction.
We will call these models $A$ and $B$, respectively.
For a given observation, model $A$ predicts $\hat{a}$ and model $B$ predicts $\hat{b}$.
When run through a SHAP explainer, we can break down these predictions even further.
Since SHAP values are additive, we know that $\hat{a} = \mu_A + s_{x_1a} + s_{x_2a}$ and $\hat{b} = \mu_B + s_{x_1b} + s_{x_2b}$.
It follows that 
\begin{align*}
  \text{avg}(\hat{a},\hat{b}) &= \frac{\hat{a}+\hat{b}}{2} \\
  &= \frac{\mu_A + s_{x_1a} + s_{x_2a} + \mu_B + s_{x_1b} + s_{x_2b}}{2} \\
  &= \frac{\mu_A + \mu_B}{2} + \frac{s_{x_1a} + s_{x_1b}}{2} + \frac{s_{x_2a} + s_{x_2b}}{2}\\
  &=\text{avg}(\mu_A, \mu_B) + \text{avg}(s_{x_1a},s_{x_1b}) + \text{avg}(s_{x_2a}, s_{x_2b}).
\end{align*}

This means that we can find the contribution to the overall model from $x_1$ by averaging $s_{x_1a}$ and $s_{x_1b}$, and likewise for the contribution to the overall model from $x_2$.

However, if we for some reason wished to stack our models such that the two outputs ($\hat{a}$ and $\hat{b}$) were multiplied, we run into a problem.
This occurs because, despite the longings of all algebra students, 
\[
\hat{a}\hat{b} = (\mu_A + s_{x_1a} + s_{x_2a})(\mu_B + s_{x_1b} + s_{x_2b}) \neq \mu_A\mu_B + s_{x_1a}s_{x_1b} + s_{x_2a}s_{x_2b}.
\]
Instead, we end up with 
\begin{align*}
    \hat{a}\hat{b} &= (\mu_A + s_{x_1a} + s_{x_2a})(\mu_B + s_{x_1b} + s_{x_2b}) \\
    &= \mu_A\mu_B + \mu_As_{x_1b} + \mu_As_{x_2b} + s_{x_1a}\mu_B + s_{x_1a}s_{x_1b} + s_{x_1a}s_{x_2b} + s_{x_2a}\mu_B + s_{x_2a}s_{x_1b} + s_{x_2a}s_{x_2b}.
\end{align*}
Even in this simple case, it is difficult to assign a single contribution to our two different variables when presented with the SHAP values of the two original models. This problem grows even more difficult with the addition of other explanatory features.
mSHAP is the methodology developed to solve this problem.

\section{The Math behind Multiplying SHAP Values}
\label{sec:math}

In a two-part model, the output of one model is multiplied by the output of a second model to obtain the response.
The principal driver behind mSHAP is the explanation of these sorts of models, and it requires that the SHAP values be multiplied together in some way to get a final SHAP value for the output.
The mathematics behind mSHAP are explained here in the general case, for any given number of predictors with a training set of arbitrary size.
Although an exact solution for the SHAP values of a two-part model is still out of reach, this method proves very accurate in its results. 

\subsection{Definitions}
\label{subsec:math2}

Consider three different models, $f, g,$ and $h$, and a single input (training) matrix $A$.
We will let the number of columns and rows in $A$ be arbitrary. 
In other words, let $A$ be an $n\times p$ matrix where each column is a covariate and each row is an observation.
Also let $A_i$ denote the $i$th observation (row) of $A$.
Furthermore, define $h$ to be the product of $f$ and $g$, so $h(A_i) = f(A_i) \cdot g(A_i)$.

Recall that the sum of the SHAP values for each covariate and the average model output must add up to the model prediction.
For simplicity in presentation, we will define $f(A_i) = \hat{x_i}$, $g(A_i) = \hat{y_i}$, and $h(A_i) = \hat{z_i}$ and the contribution of the $j$th predictor to $x_i$ as $s_{x_ij}$.
With these considerations in place, we can define the output space of our three models on the training data set.

For model $f$:
\begin{align*}
\hat{x_1} = & s_{x_11} + s_{x_12} + s_{x_13} + \ldots + s_{x_1p} + \mu_f \\
\hat{x_2} = &s_{x_21} + s_{x_22} + s_{x_23} + \ldots + s_{x_2p} + \mu_f \\
\hat{x_3} = & s_{x_31} + s_{x_32} + s_{x_33} + \ldots + s_{x_3p} + \mu_f \\
\vdots &\\
\hat{x_n} = & s_{x_n1} + s_{x_n2} + s_{x_n3} + \ldots + s_{x_np} + \mu_f
\end{align*}

For model $g$:
\begin{align*}
\hat{y_1} = & s_{y_11} + s_{y_12} + s_{y_13} + \ldots + s_{y_1p} + \mu_g \\
\hat{y_2} = &s_{y_21} + s_{y_22} + s_{y_23} + \ldots + s_{y_2p} + \mu_g \\
\hat{y_3} = & s_{y_31} + s_{y_32} + s_{y_33} + \ldots + s_{y_3p} + \mu_g \\
\vdots &\\
\hat{y_n} = & s_{y_n1} + s_{y_n2} + s_{y_n3} + \ldots + s_{y_np} + \mu_g
\end{align*}

And for model $h$:
\begin{align*}
\hat{z_1} = & s_{z_11} + s_{z_12} + s_{z_13} + \ldots + s_{z_1p} + \mu_h \\
\hat{z_2} = &s_{z_21} + s_{z_22} + s_{z_23} + \ldots + s_{z_2p} + \mu_h \\
\hat{z_3} = & s_{z_31} + s_{z_32} + s_{z_33} + \ldots + s_{z_3p} + \mu_h \\
\vdots &\\
\hat{z_n} = & s_{z_n1} + s_{z_n2} + s_{z_n3} + \ldots + s_{z_np} + \mu_h
\end{align*}

Furthermore, given our training data $A$, we can extract the values of $\mu_f, \mu_g,$ and $\mu_h$.
As explained above, these are the average value of the model predictions on the training set.

\begin{align*}
\mu_f &= \frac{1}{n}\sum_{i = 1}^n \hat{x_i} = \frac{\hat{x_1} + \hat{x_2} + \hat{x_3} + \ldots + \hat{x_n}}{n} \\
\mu_g &= \frac{1}{n}\sum_{i = 1}^n \hat{y_i} = \frac{\hat{y_1} + \hat{y_2} + \hat{y_3} + \ldots + \hat{y_n}}{n} \\
\mu_h &= \frac{1}{n}\sum_{i = 1}^n \hat{z_i} = \frac{\hat{z_1} + \hat{z_2} + \hat{z_3} + \ldots + \hat{z_n}}{n}
\end{align*}

In practice, it is necessary to be able to pull $\mu_h$ out of $\hat{x_i}\hat{y_i}$.
When implemented, it is important to note that $\mu_f\mu_g = \mu_f\mu_g - \mu_h + \mu_h$.
Since every expansion of SHAP values from $\hat{x_i}\hat{x_j}$ contains $\mu_f\mu_g$, we substitute in $\mu_f\mu_g - \mu_h + \mu_h$, where $\mu_h$ is essential and $\mu_f\mu_g - \mu_h$ becomes a term that we label $\alpha$ and distribute among all the SHAP values.
A more formalized definition of $\alpha$ is given in \ref{methods:relationship}.

\subsection{Obtaining $z_i$'s SHAP Values}
\label{subsec:math3}

We now derive the individual SHAP values for each variable as it pertains to the prediction of model $h$.
Again, we will let this output be an arbitrary $\hat{z_i}$.
Recall that
\[
\hat{z_i} = \hat{x_i}\hat{y_i} = (s_{x_i1} + s_{x_i2} + s_{x_i3} + \ldots + s_{x_ip} + \mu_f)(s_{y_i1} + s_{y_i2} + s_{y_i3} + \ldots + s_{y_ip} + \mu_g).
\]
Using a tabular form for visual simplicity, we obtain the following expansion:

\begin{displaymath}
\begin{array}{c|ccccccccccc}
&s_{x_i1} &+& s_{x_i2} &+& s_{x_i3} &+& \ldots &+& s_{x_ip} &+& \mu_f \\[6pt]
\hline \\
s_{y_i1}&  s_{x_i1} s_{y_i1}&& s_{x_i2} s_{y_i1}&& s_{x_i3}s_{y_i1} && \ldots && s_{x_ip} s_{y_i1}&& \mu_f s_{y_i1} \\
+ \\
s_{y_i2} &s_{x_i1} s_{y_i2} && s_{x_i2}s_{y_i2}  && s_{x_i3} s_{y_i2} && \ldots && s_{x_ip} s_{y_i2} && \mu_f s_{y_i2} \\
+ \\
s_{y_i3} &s_{x_i1}s_{y_i3} && s_{x_i2}s_{y_i3} && s_{x_i3} s_{y_i3}&& \ldots && s_{x_ip} s_{y_i3}&& \mu_f s_{y_i3} \\
+ \\
\vdots & \vdots && \vdots & &\vdots && \ddots &&\vdots && \vdots\\
+ \\
s_{y_in} & s_{x_i1}s_{y_ip}  && s_{x_i2}s_{y_ip}  && s_{x_i3}s_{y_ip}  && \ldots && s_{x_ip}s_{y_ip}  && \mu_f s_{y_ip} \\
+ \\
\mu_g & s_{x_i1}\mu_g  && s_{x_i2}\mu_g && s_{x_i3}\mu_g  && \ldots && s_{x_ip}\mu_g  && \mu_f\mu_g 
\end{array}
\end{displaymath}

We break these terms into the SHAP values for each variable, 1 through $p$, for $\hat{z_i}$.
Our approach breaks $s_{z_ij}$ into two parts, which we call $s_{z_ij}'$ and $\alpha_{ij}$.
Though the method of obtaining $\alpha_i$ can take on several forms, $s_{z_ij}'$ is always as follows (where $j$ refers to the $j$th covariate):
\begin{align*}
s_{z_ij}' &= \mu_fs_{y_ij} + s_{x_ij}\mu_g + s_{x_ij}s_{y_ij} + \sum^p_{a = 1}(\frac{s_{x_ij}s_{y_ia}}{2}I(a\neq j)) + \sum^p_{a = 1}(\frac{s_{y_ij}s_{x_ia}}{2}I(a\neq j)) \\
&= \mu_fs_{y_ij} + s_{x_ij}\mu_g + \frac{1}{2}\sum^p_{a = 1}(s_{x_ij}s_{y_ia} + s_{y_ij}s_{x_ia})
\end{align*}
In words, and with the aid of the table above, this can be described as the sum of the $i$th row and $i$th column, where every term is divided by two except the terms with $\mu_f$ and $\mu_g$.
When applied to each variable, this can be written as 
\[
\hat{z_i} = \sum^p_{j = 1}\left[\mu_fs_{y_ij} + s_{x_ij}\mu_g + \frac{1}{2}\sum^p_{a = 1}(s_{x_ij}s_{y_ia} + s_{y_ij}s_{x_ia})\right] + \mu_f\mu_g
\]
and applying the breakdown we derived above, while simplifying as well,
\[
\hat{z_i} = \left(\sum^p_{j = 1}s_{z_ij}'\right) + \alpha + \mu_h.
\]

For a proof that this formula and the subsequent distribution of $\alpha$ maintains the local accuracy property of SHAP values, see \ref{methods:proof}.

\subsection{Methods for distributing $\alpha$}
\label{subsec:math4}

We now arrive at the aforementioned point of deciding how to distribute $\alpha$ into each $s_{z_ij}$.
There are four ways which we tested for distributing $\alpha$, the first being simple uniform distribution, and the others being variations of weighting based on the value of $s_{z_ij}'$.
Note that all four of these methods maintain the local accuracy property of SHAP values.

\subsubsection{Uniformly Distributed}

The simplest way of distributing $\alpha$ between all the $s_{z_ij}$'s is to divide it evenly.
In this case, our resulting equation for each variable's SHAP value would be 
\[
s_{z_ij}= s_{z_ij}'+ \frac{\alpha}{p}.
\]
This method could prove a strong baseline.

\subsubsection{Raw Weights}
The computation of this method is made easier by recalling that 
\[
\sum^p_{j = 1}s_{z_ij}' = \hat{z_i} - \mu_f\mu_g,
\]
which allows us to use $\hat{z_i} - \mu_f\mu_g$ as the whole upon which we base our weighting.
When applied, this method defines each SHAP value as
\[
s_{z_ij}= s_{z_ij}'+ \frac{s_{z_ij}'}{\hat{z_i} - \mu_f\mu_g}(\alpha).
\]

\subsubsection{Absolute Weights}
This method differs from that of the raw weights, in that instead of summing the $s_{z_ij}'$'s we sum their absolute values.
The weight for each SHAP value is calculated with
\[
s_{z_ij}= s_{z_ij}'+ \frac{|s_{z_ij}'|}{\sum^p_{k = 1}|s_{z_ik}'|}(\alpha).
\]

\subsubsection{Squared Weights}
Finally, instead of working with the absolute values, we could work with squares.
Similar to the equation above, the SHAP values under this method are computed by
\[
s_{z_ij}= s_{z_ij}'+ \frac{(s_{z_ij}')^2}{\sum^p_{k = 1}(s_{z_ik}')^2}(\alpha).
\]

\section{Simulation Study for Distributing $\alpha$}
\label{sec:sim}

To test the differences between these methods of distributing $\alpha$, we simulated various multiplicative models based on known equations, and compared the results of our multiplicative method with the output from kernelSHAP.
KernelSHAP is an existing generalized method for estimating the contributions based on any prediction function.
However, it is extremely computationally expensive when compared with TreeSHAP.
When training on millions of rows with many variables, it becomes unrealistic to use kernelSHAP for computing the SHAP values.

\subsection{Scoring the Methods}
\label{subsec:sim1}

Several factors were considered in scoring, including the mean absolute error of the SHAP values, the directions of the SHAP values, and the rank (in magnitude) of the SHAP values for each variable.
The score needed to be a singular way to asses how close the method gets to the kernelSHAP estimates.
Even though kernelSHAP is an estimate and not necessarily the truth, we used it as a benchmark in the different parts of our score.
This allowed us to compare the new variations of the mSHAP method to existing methods for the computation of SHAP values.

For ease of notation, if we define the SHAP value we are estimating as $s_{z_ij}$, then we can define its counterpart as computed by kernelSHAP, as $k_{z_ij}$.

\subsubsection{General Equation for Scoring}

In the end, an equation was formed to create a raw ``score'' based on the direction of the SHAP value, the relative value of the SHAP value, and the rank (importance) of the SHAP value in comparison to kernelSHAP.
The score ranges from 0 to 3 (with 3 being the best possible score), and is defined by
\begin{align*}
\beta(s_{z_ij},k_{z_ij} | \theta_1, \theta_2) &= \lambda_1(s_{z_ij}, k_{z_ij} | \theta_1) + \lambda_2(s_{z_ij},k_{z_ij} | \theta_2) + \lambda_3(s_{z_ij},k_{z_ij})
\end{align*}
where
\begin{align*}
\lambda_1(s_{z_ij},k_{z_ij}|\theta_1) &= \begin{cases}
					  1 & \quad s_{z_ij}k_{z_ij} > 0 \\
					  \text{min}(1, \frac{1 + \theta_1}{|s_{z_ij}| + |k_{z_ij}| + \theta_1}) &\quad\text{otherwise} \\
					  \end{cases} \\
\lambda_2(s_{z_ij},k_{z_ij}|\theta_2) &=\text{min}(1, \frac{1 + \theta_2}{|s_{z_ij} - k_{z_ij}| + 1}) \\
\lambda_3(s_{z_ij},k_{z_ij}) &= \frac{1}{|\text{imp}(s_{z_ij}) - \text{imp}(k_{z_ij})| + 1} \\
\end{align*}
and imp$(s_{z_ij})$ is the importance of that SHAP value relative to the other contributions in the observation (where importance is determined by absolute value).

In this function (and as will be described in the following section), $\lambda_1$ is the contribution from the signs of the SHAP values, $\lambda_2$ is the contribution from the relative value of the SHAP values, and $\lambda_3$ is the contribution from the relative ranking (importance) of the SHAP values.

\subsubsection{Lambda Functions}

\begin{figure}[h!]
  \centering
  \begin{subfigure}[b]{0.4\linewidth}
    \includegraphics[width=\linewidth]{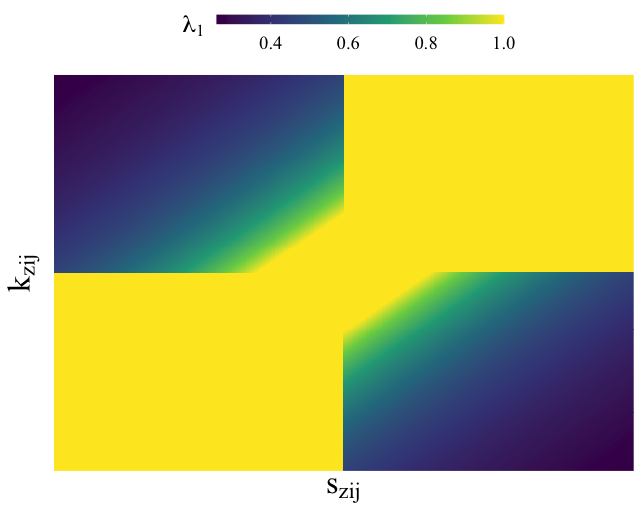}
    \caption{$\lambda_1$}
  \end{subfigure}
  \begin{subfigure}[b]{0.4\linewidth}
    \includegraphics[width=\linewidth]{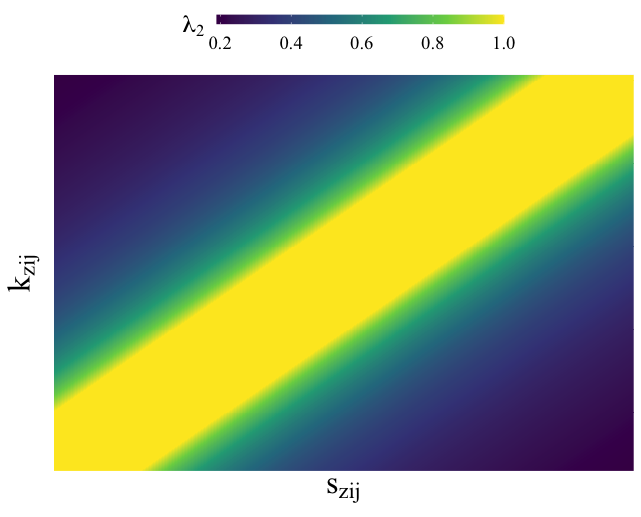}
    \caption{$\lambda_2$}
  \end{subfigure}
  \caption{Heat maps for the $\lambda$ functions.}
  \label{fig:lambda}
\end{figure}

To gain some intuition about the $\lambda$ functions and the impact of $\theta_1$ and $\theta_2$ we depict them in Figure \ref{fig:lambda}.

For $\lambda_1$, which measures whether the two SHAP values are the same sign, any values in the first and third quadrants return a perfect score of 1, since the two values have the same sign.
It also allows for some wiggle room with $\theta_1$, by allowing anything within the lines $k_{z_ij} = s_{z_ij} + \theta_1$ and $k_{z_ij} = s_{z_ij} - \theta_1$ to be 1.  Beyond those boundaries, the scores gradually decrease.

The function $\lambda_2$, which compares the values, also creates boundary lines for the perfect score of 1 at $k_{z_ij} = s_{z_ij} + \theta_2$ and $k_{z_ij} = s_{z_ij} - \theta_2$.
In other words, as long as the difference between $s_{z_ij}$ and $k_{z_ij}$ is less than $\theta_2$, the function will return 1.
Beyond that and the value begins to decrease.

Out of the three $\lambda_3$, the rank measure, is the easiest to understand.
In a given observation, each SHAP value is given a rank (between 1 and $p$, inclusive) based on its absolute value.
These ranks are then compared, and the closer they are together, the higher the score, with a perfect score of 1 being obtained if the two rankings are the same.

\subsection{Simulation Study}
\label{subsec:sim2}

As mentioned above, we simulated various multiplicative models based on known equations, and compared the results of our multiplicative method with the output from kernelSHAP in order to test the model.

Specifically, we used three variables, $x_1, x_2, x_3$ in a variety of response equations $y_1, y_2$ to create models for $y_1$ and $y_2$ and then multiply their outputs together.
Using the multiplied output and the covariates, we were able to use kernelSHAP to compute an estimate of the SHAP values.
We could then compare this estimate to the result from our multiplicative method, as described above, with different ways of distributing $\alpha$ applied.

More details on the simulation can be found in \ref{methods:simulation}.

For testing, we used 100 samples in each iteration for faster computation, which allowed us to simulate over 2500 scenarios.
Specifically, we worked with all possible combinations of the following values.

\begin{center}
\begin{tabular}{c c}
Variable & Possible Values \\
\toprule
$y_1$ & $x_1 + x_2 + x_3$ \\
     & $2*x_1 + 2*x_2 + 3*x_3 $\\
\midrule
$y_2$ & $x_1 + x_2 + x_3$ \\
    & $2*x_1 + 2*x_2 + 3*x_3$ \\
    & $x_1 * x_2 * x_3$ \\
    & $x_1^2 * x_2^3 * x_3^4 $\\
    & $(x_1 + x_2) / (x_1 + x_2 + x_3)$ \\
    & $x_1 * x_2 / (x_1 + x_1 * x_2 + x_1^2 * x_3^2)  $\\
 \midrule
 $\theta_1$ & 1.5, 2.5, 3.5, 4.5, 5.5, 6.5, 7.5, 8.5, 9.5, 10.5, \\
 & 11.5, 12.5, 13.5, 14.5, 15.5, 16.5, 17.5, 18.5, 19.5, 20.5 \\
 \midrule
 $\theta_2$ & 1, 6, 11, 16, 21, 26, 31, 36, 41, 46\\
 \midrule
\end{tabular}
\end{center}

For each combination of values in the above table, we distributed $\alpha$ in each of the four ways mentioned in Section \ref{subsec:math4}.
The resulting table, therefore, had results for each model and each way of distributing $\alpha$. 
In general, we averaged across all rows of the same method to obtain the scores that were compared to each other.

It should also be noted that in our examples, our covariates were distributed as follows:
\begin{align*}
x_1 &\sim \text{Uniform} [-10, 10] \\
x_2 &\sim \text{Uniform} [0, 20] \\
x_3 &\sim \text{Uniform} [-5, -1].
\end{align*}

\subsection{Results of the Simulation}
\label{subsec:sim3}

In general, the multiplicative SHAP method did very well, when compared to the kernelSHAP output.
Since kernelSHAP is an estimation as well, it is hard to determine exactly how well the multiplicative SHAP method does, but we will summarize some statistics here.

\subsubsection{Distributing $\alpha$}

After trying the aforementioned four methods for distributing $\alpha$ into the SHAP values, we came to the conclusion that the weighted by absolute value method was the best.
This came by way of the score as well as other metrics.
Details can be seen in the following table (please note that all values are averaged across all 2520 simulations).

{\small
\begin{center}
\begin{tabular}{c|r r r r r r}
Method & Score & \makecell{Direction\\Score} & \makecell{Relative\\Value Score} &  Rank Score & \makecell{Pct Same\\Sign} & \makecell{Pct Same\\Rank}\\
\hline
&&&\\
\makecell{Weighted by\\Absolute Value} & \textbf{2.27} & \textbf{0.869} & \textbf{0.594} & \textbf{0.802} & \textbf{84.8\%} & \textbf{62.5\%} \\
&&&\\
\makecell{Weighted by\\Squared Value} & 2.21 & 0.841 & 0.579 & 0.792 & 81.8\% & 60.8\% \\
&&&\\
\makecell{Uniformly\\Distributed} & 2.20 & 0.858 & 0.563 & 0.783 & 83.7\% & 59.4\% \\
&&&\\
\makecell{Weighted by\\Raw Value} & 1.99 & 0.727 & 0.494 & 0.768 & 71.4\% & 56.2\% \\
\end{tabular}
\end{center}
}

\subsubsection{Impact of $\theta_1$ and $\theta_2$}

We plotted the effects of the different values for $\theta_1$ and $\theta_2$ on the overall score, based on type of method of distribution.

\begin{figure}
    \centering
    \includegraphics[width=13cm]{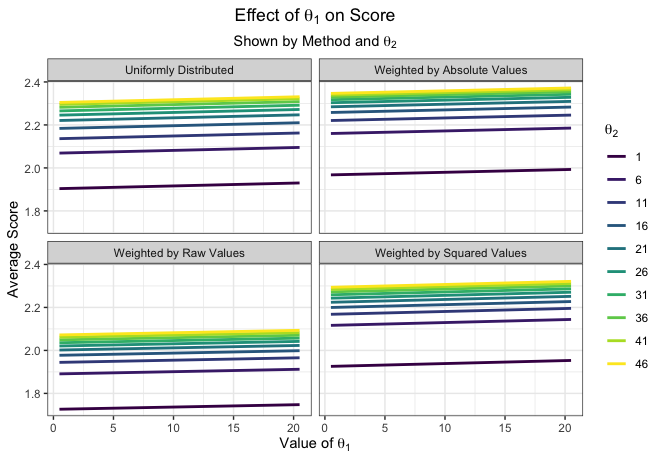}
    \caption{How $\theta_1$ impacts overall score, on average}
    \label{fig:theta_1}
\end{figure}

\begin{figure}
    \centering
    \includegraphics[width=13cm]{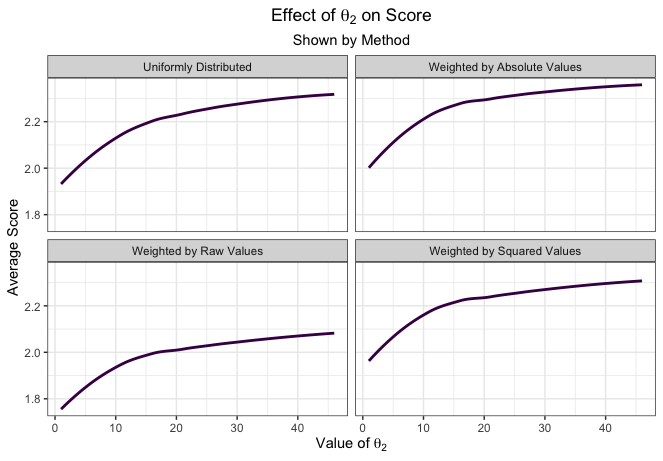}
    \caption{How $\theta_2$ impacts overall score, on average}
    \label{fig:theta_2}
\end{figure}

As can be seen, in Figures \ref{fig:theta_1} and \ref{fig:theta_2}, changing the value of these two parameters has a similar impact across all scoring methods.

\subsubsection{Computational Time}

The most dramatic benefit of mSHAP over kernelSHAP is the computational efficiency of mSHAP. 

\begin{figure}[h!]
  \centering
  \begin{subfigure}[b]{0.4\linewidth}
    \includegraphics[width=\linewidth]{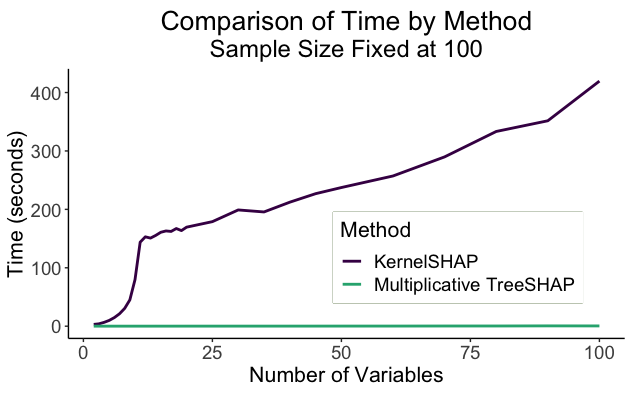}
    \caption{Fixed $n$}
  \end{subfigure}
  \begin{subfigure}[b]{0.4\linewidth}
    \includegraphics[width=\linewidth]{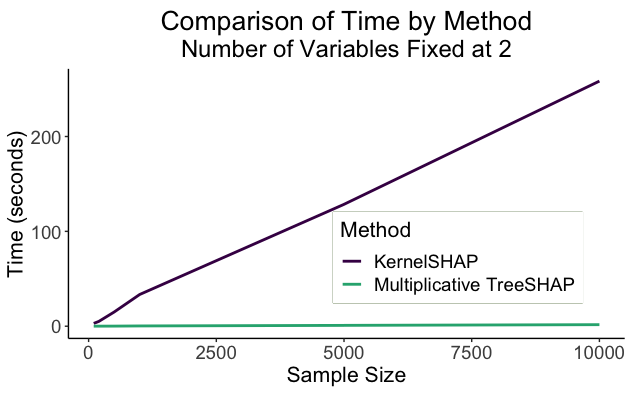}
    \caption{Fixed Variables}
  \end{subfigure}
  \caption{Computational time of kernelSHAP and mSHAP}
  \label{fig:times}
\end{figure}

In Figure \ref{fig:times}, we are able to see the comparison in run time between kernelSHAP and the mSHAP method (including the individual treeSHAP value calculations).
Both increasing the number of variables and the number of samples causes the time of kernelSHAP to grow greatly, while the multiplicative method remains fairly constant.
In these trials, the number of background samples was fixed at 100 for kernelSHAP.

A case study can show the importance of this.
In the auto insurance data set, there are 5,000,000 rows in the test set, with 46 variables.
For the sake of simplicity, let's assume that we use 45 of those variables, and that 100 background samples is enough to compute accurate SHAP values.
In reality, it would need many more background samples, but that only accentuates the point, as a large quantity of  background samples slows  kernelSHAP drastically.
KernelSHAP computes SHAP values for 45 variables at a rate of about 2.268 seconds per observation on a personal laptop.
In order to compute the SHAP values for the entire test set, then, one would need about 131 days of continuous compute time.

In contrast, our multiplicative method, using treeSHAP on two tree-based models, computes SHAP values at a rate of about .00175 seconds per observation for a model with 45 variables.
To compute the SHAP values for the entire test set using this method would take a little less than three hours of continuous compute time.

\subsection{Final Equation for mSHAP}

Based on the results of the simulation, we determine that the best way of distributing $\alpha$ is the method of weighting by absolute values (as described above).
Recall from above that in this method,
\[
s_{z_ij}= s_{z_ij}'+ \frac{|s_{z_ij}'|}{\sum^p_{k = 1}|s_{z_ik}'|}(\alpha)
\]
and that $s_{z_ij}'$ refers to an initial mSHAP value, before the correction introduced by $\alpha$.  It is calculated as
\[
s_{z_ij}' = \mu_fs_{y_ij} + s_{x_ij}\mu_g + \frac{1}{2}\sum^p_{a = 1}(s_{x_ij}s_{y_ia} + s_{y_ij}s_{x_ia}).
\]

Thus, the final equation for the mSHAP value of the $j$th predictor on the $i$th observation can be written as
\[
s_{z_ij} = \mu_fs_{y_ij} + s_{x_ij}\mu_g + \frac{1}{2}\left[\sum^p_{a = 1}(s_{x_ij}s_{y_ia} + s_{y_ij}s_{x_ia})\right] + \frac{|s_{z_ij}'|}{\sum^p_{k = 1}|s_{z_ik}'|}(\alpha).
\]
For a complete proof that local accuracy holds with this equation, see \ref{methods:proof}.

\section{Case Study}
\label{sec:app}

In order to prove the efficacy of mSHAP it is necessary to put it into practice.
We obtained an insurance dataset including over 20 million auto insurance policies for a large insurance provider in the United States.
Using this data, we created a two-part model that predicts the expected property damage cost of each policy.
Both parts of this model consist of tree-based methods, specifically random forests.
After creating this model, we used the shap python library to explain the predictions of each individual part on a sample of 50,000 observations from our test set.
We then applied the final mSHAP method as described above, to obtain explanations for the overall model and used the mshap R package to visualize some of the results.

\subsection{Model Creation}
\label{subsec:app1}

As mentioned above, the model is a two-part model for predicting the expected cost of the policy.
The first part of the model predicts the frequency of the claims.
It is a random forest that predicts the probability of each of 4 possible outcomes (a multinomial model).
In our data set, there existed policies with up to 7 claims, but we chose the classes of 0, 1, 2, and 3 and bundled everything over 3 into the 3 class.
The data was heavily imbalanced, so we used a combination of upsampling the minority classes (1, 2, 3 claims) and downsampling the majority class (0 claims) to get a more balanced training data set.
This allowed the model to use the information to predict meaningful probabilities instead of always assigning a very high probability to 0 claims.

The second part is a random forest which predicts the severity component of the two-part model, or the expected cost per claim.

Once these models were created, we could calculate the expected value (or in this case, the expected cost) of a policy in the following manner.
If we let $\hat{P_i}(a)$ denote the predicted probability of the for the $i$th policy of the $a$th class and $\hat{y_i}$ be the predicted severity of the policy, then
\[
EV = \hat{y_i} \left(0\hat{P_i}(0) + 1\hat{P_i}(1) + 2\hat{P_i}(2) + 3\hat{P_i}(3)\right)
\]

The final two-part model was used to predict the expected cost of 50,000 policies from the test data set.
For more specific details about the model and how it was tuned, see \ref{methods:model}.

\subsection{Model Explanation}
\label{subsec:app2}

\begin{figure}[h!]
  \centering
  \begin{subfigure}[b]{0.8\linewidth}
    \includegraphics[width=\linewidth]{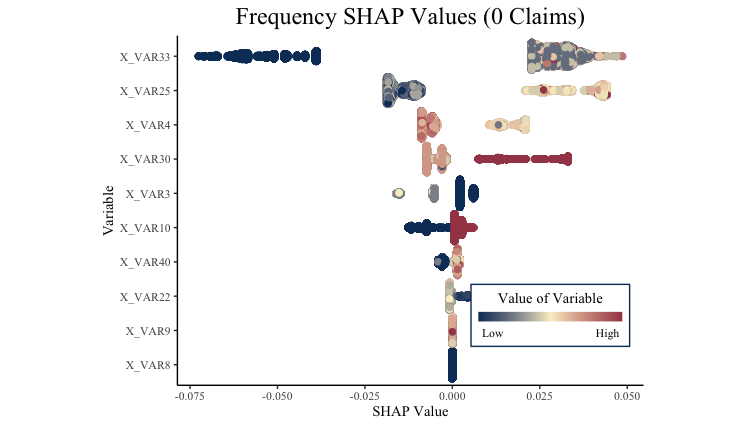}
    \caption{Summary plot of the frequency model's SHAP values for the 0 claim class}
  \end{subfigure}
  \begin{subfigure}[b]{0.8\linewidth}
    \includegraphics[width=\linewidth]{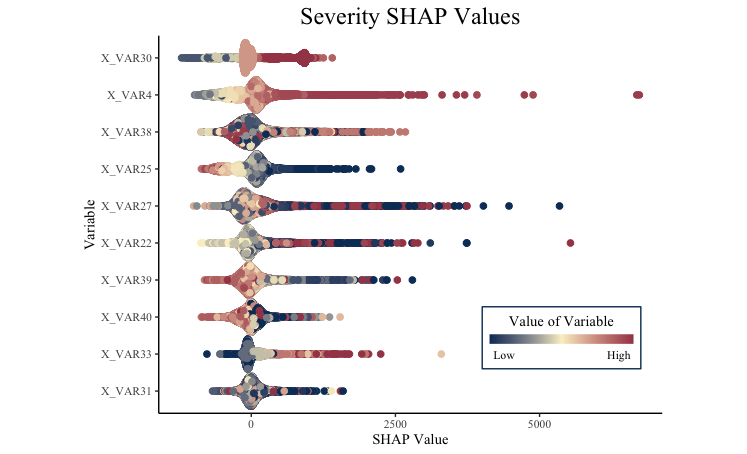}
    \caption{Summary plot of the severity model's SHAP values}
  \end{subfigure}
  \caption{Example summary plots of SHAP values from the individual model parts}
  \label{fig:shap_parts}
\end{figure}

After creating the two-part model and getting final predictions for the expected cost of the claims, we were able to apply mSHAP to explain the final model predictions.
Before doing this, we computed the SHAP values on the individual models so that we have the necessary data to apply the mSHAP method for explaining two-part models.
Summary plots for the five different sets of SHAP values (one for severity, and one for each class of the frequency model) can be created.
In Figure \ref{fig:shap_parts} we depict the SHAP values for one of the frequency classes from the frequency model and the SHAP values for the severity model.

\begin{figure}
    \centering
    \includegraphics[width=13cm]{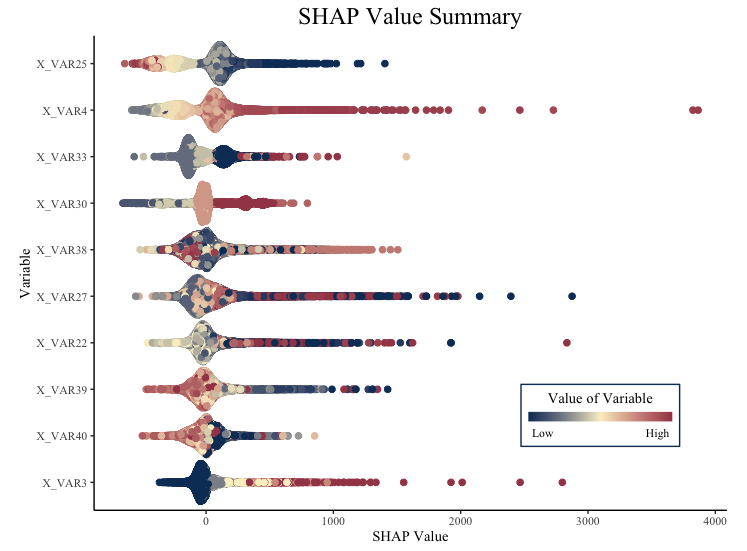}
    \caption{Summary plot of the two-part model's mSHAP values.}
    \label{fig:mshap_sum}
\end{figure}

\begin{figure}
    \centering
    \includegraphics[width=13cm]{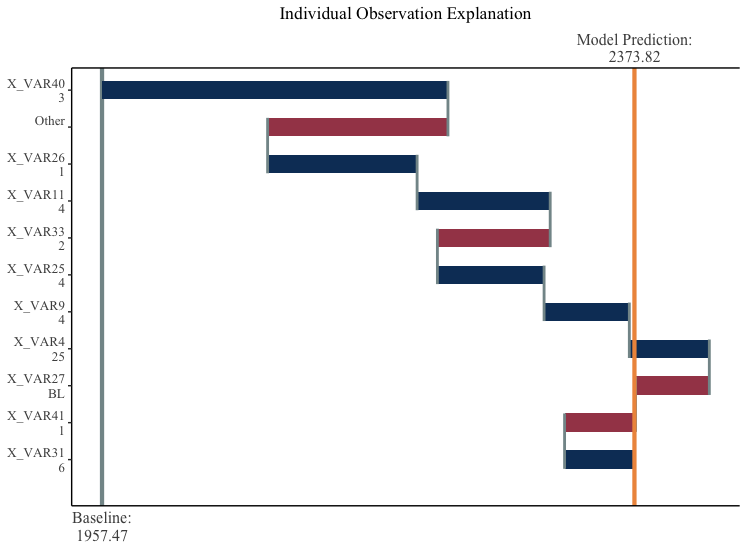}
    \caption{Observation plot from the two-part model's mSHAP values. This plot shows how mSHAP can be used to explain a single observation.}
    \label{fig:mshap_obs}
\end{figure}

After computing these SHAP values, we applied the mSHAP method detailed in this paper.
Note that after applying mSHAP, the expected value formula above is simply a linear combination, and we are able to perform that same linear combination on the SHAP values before (or after) applying mSHAP.
This process left us with a single mSHAP value for each variable in every row of our test set, and an overall expected value across the training set.
The summary plot of those final mSHAP values can be seen in Figure \ref{fig:mshap_sum}, and an example of an observation plot is shown in Figure \ref{fig:mshap_obs}.

The beauty of the mSHAP method is it allows for a two-part model to be explained in the same ways that tree-based models can be easily explained with SHAP values.
As can be seen in the plots, general trends across variables can be established, as well as specific policies dissected to see the individual motivators behind each prediction.
The ability of mSHAP to explain these types of models opens the door to using two-part models that are both powerful and explainable.

\section{Conclusion}
\label{sec:conc}

In this paper we developed mSHAP, a method for calculating SHAP values in two-part models. Our method is much quicker than kernelSHAP and may lead to increased adoption of machine learning (especially tree-based) components of two-part models. This is particularly true in insurance where issues of fairness and discrimination are very important.

\appendix

\section{Shapley Values}
\label{methods:shapley_vals}

In this section we briefly discuss the math behind Shapley values.
Please note that this section leans heavily upon the explanations and formulas as given in \citet{NIPS2017_7062}.
A motivated reader will find further information regarding Shapley values in that paper.

Shapley values are a class of what is known as additive feature attribution methods.
These methods are defined as methods that have an ``explanation model that is a linear function of binary variables:''
\[
g(z') = \phi_0 + \sum^M_{i = 1}\phi_iz'_i
\]
where $M$ is the number of input features, $\phi_0,\phi_i\in\mathbb{R}$ and $z'_i\in \{0, 1\}^M$.
Essentially, every prediction of the model (which we will denote $f(x)$), can be obtained by assigning some contribution to each of the variables.

The Shapley values have three desirable properties, as mentioned above, and the formal definitions for these properties are given here.

\textbf{Local Accuracy}.
Local accuracy requires that the outputs of our model $f(x)$ and the outputs of the additive feature attribution method to be equal.
In symbols, this means that
\[
f(x) = g(x') = \phi_0 + \sum^M_{i = 1}\phi_ix'_i.
\]

\textbf{Missingness}.
A second property is missingness.
Simply stated, any variable that has a value of 0 requires its corresponding contribution to the output to be zero.
In other words, 
\[
x'_i = 0 \Rightarrow \phi_i = 0
\]

\textbf{Consistency}.
The third property is consistency, which assures that if the model changes so that an input's contribution increases or stays the same, the attribution of that input should not decrease.
If we let $f_x(z') = f(h_x(z'))$ and $z'/i$ denote setting $z_i' = 0$, then for any two models $f$ and $f'$, if
\[
f'_x(z') - f'_x(z'/i) \geq f_x(z') - f_x(z'/i)
\]
for all inputs $z'\in \{0,1\}^M$, then $\phi_i(f', x) \geq \phi_i(f,x)$.

The theorem proposed by \citeauthor{NIPS2017_7062} is as follows: Only one possible explanation model follows the above definition and the three given properties.
\[
\phi_i(f,x) = \sum_{z'\subseteq x'}\frac{|z'|!(M - |z'| - 1)!}{M!}[f_x(z') - f_x(z' / i)]
\]
where $|z'|$ is the number of non-zero entries in $z'$ and $z' \subseteq x'$ represents all $z'$ vectors where the non-zero entries are asubset of the non-zero entries in $x'$.

\section{The Relationship Between $\mu_f, \mu_g$ and $\mu_h$}
\label{methods:relationship}

Recall from above that 
\[
\mu_h = \frac{1}{n}\sum_{i = 1}^n \hat{z_i} = \frac{\hat{z_1} + \hat{z_2} + \hat{z_3} + \ldots + \hat{z_n}}{n},
\]
and that we defined model $h$ as the product of models $f$ and $g$.
Thus, any $\hat{z_i}$ is equivalent to $\hat{x_i}\hat{y_i}$.

Taking the equation above and substituting $\hat{x_i}\hat{y_i}$ for every $\hat{z_i}$ we see that 
\[
\mu_h = \frac{\hat{x_1}\hat{y_1} + \hat{x_2}\hat{y_2} + \hat{x_3}\hat{y_3} + \ldots + \hat{x_n}\hat{y_n}}{n}.
\]

Whenever we multiply $\hat{x_i}$ and $\hat{y_i}$ to get $\hat{z_i}$, it is inevitable that we end up with the term $\mu_f\mu_g$ in the resulting expansion.
We will take this term and split it into two parts: $\mu_h$ and $\alpha$, some correction that must be added in to the other SHAP values.
Start with the expansion of $\mu_f\mu_g$,
\[
\mu_f\mu_g = (\frac{1}{n}\sum_{i = 1}^n \hat{x_i} ) \cdot (\frac{1}{n}\sum_{i = 1}^n \hat{y_i} )
\]
which can be written in tabular form for ease of explanation.
\begin{displaymath}
\begin{array}{c|ccccccccc}
& \frac{\hat{x_1}}{n} &+& \frac{\hat{x_2}}{n} &+&\frac{\hat{x_3}}{n} &+& \ldots &+& \frac{\hat{x_n}}{n} \\[6pt]
\hline \\
 \frac{\hat{y_1}}{n} &  \frac{\hat{x_1}\hat{y_1}}{n^2} & & \frac{\hat{x_2}\hat{y_1}}{n^2} & & \frac{\hat{x_3}\hat{y_1}}{n^2} & & \ldots & & \frac{\hat{x_n}\hat{y_1}}{n^2} \\
 +&&&&&&&&& \\
 \frac{\hat{y_2}}{n} & \frac{\hat{x_1}\hat{y_2}}{n^2} && \frac{\hat{x_2}\hat{y_2}}{n^2} && \frac{\hat{x_3}\hat{y_2}}{n^2} & &\ldots && \frac{\hat{x_n}\hat{y_2}}{n^2} \\
  +&&&&&&&&& \\
 \frac{\hat{y_3}}{n}  &\frac{\hat{x_1}\hat{y_3}}{n^2} & &\frac{\hat{x_2}\hat{y_3}}{n^2} & &\frac{\hat{x_3}\hat{y_3}}{n^2} && \ldots && \frac{\hat{x_n}\hat{y_3}}{n^2} \\
  +&&&&&&&&& \\
 \vdots & \vdots && \vdots & &\vdots && \ddots &&\vdots \\
  +&&&&&&&&& \\
 \frac{\hat{y_n}}{n} & \frac{\hat{x_1}\hat{y_n}}{n^2} && \frac{\hat{x_2}\hat{y_n}}{n^2} && \frac{\hat{x_3}\hat{y_n}}{n^2} && \ldots && \frac{\hat{x_n}\hat{y_n}}{n^2} \\
\end{array}
\end{displaymath}

Note that along the diagonal are the terms that may be of interest to us, specifically
\[
\sum^n_{i = 1}\frac{\hat{x_i}\hat{y_i}}{n^2} = \frac{\mu_h}{n}.
\]

By multiplying both sides by $n$, we see that 
\[
n\sum^n_{i = 1}\frac{\hat{x_i}\hat{y_i}}{n^2} = \mu_h.
\]
And since we already have one, we can simply add $n - 1$ and subtract $n - 1$ summands to pull out the desired $\mu_h$.
This can be summarized as follows
\begin{align*}
\mu_f\mu_g &= \sum^n_{i = 1}\sum^n_{j = 1} (\frac{\hat{x_i}\hat{y_j}}{n^2}I(i\neq j)) - (n-1)\sum^n_{i = 1}\frac{\hat{x_i}\hat{y_i}}{n^2} + \sum^n_{i = 1}\frac{\hat{x_i}\hat{y_i}}{n} \\
&= \sum^n_{i = 1}\sum^n_{j = 1} (\frac{\hat{x_i}\hat{y_j}}{n^2}I(i\neq j)) - (n-1)\sum^n_{i = 1}\frac{\hat{x_i}\hat{y_i}}{n^2} + \mu_h \\
&= \alpha + \mu_h
\end{align*}
where $\alpha = \sum^n_{i = 1}\sum^n_{j = 1} (\frac{\hat{x_i}\hat{y_j}}{n^2}I(i\neq j)) - (n-1)\sum^n_{i = 1}\frac{\hat{x_i}\hat{y_i}}{n^2} = \mu_f\mu_g - \mu_h$.
This becomes a critical element in our substitutions in later steps.

\subsection{Proof of Local Accuracy}
\label{methods:proof} 

If we define $\hat{z_i}$ as the prediction of our model, $h$ for the $i$th observation, $\mu_h$ as the average model prediction across our training set, and $s_{z_ij}$ as the contribution of the $j$th variable to the $i$th observation's prediction, we can define local accuracy as
\[
\hat{z_i} = \mu_h + \sum^p_{j = 1}s_{z_ij}.
\]
In this section we will prove that this equation holds for our chosen definition of $s_{z_ij}$.

Remember that based on our initial definition, $\hat{z_i} = \hat{x_i}\hat{y_i}$,
and recall from above the final equation for the mSHAP values:
\[
s_{z_ij} = \mu_fs_{y_ij} + s_{x_ij}\mu_g + \frac{1}{2}\left[\sum^p_{a = 1}(s_{x_ij}s_{y_ia} + s_{y_ij}s_{x_ia})\right] + \frac{|s_{z_ij}'|}{\sum^p_{k = 1}|s_{z_ik}'|}(\alpha).
\]
We see that 
\begin{align*}
\mu_h + \sum^p_{j = 1}s_{z_ij} &= \mu_h + \sum^p_{j = 1}\left(\mu_fs_{y_ij} + s_{x_ij}\mu_g + \frac{1}{2}\left[\sum^p_{a = 1}(s_{x_ij}s_{y_ia} + s_{y_ij}s_{x_ia})\right] + \frac{|s_{z_ij}'|}{\sum^p_{k = 1}|s_{z_ik}'|}(\alpha) \right) \\
&= \mu_h + \left(\mu_fs_{y_i1} + s_{x_i1}\mu_g + \frac{1}{2}\left[\sum^p_{a = 1}(s_{x_i1}s_{y_ia} + s_{y_i1}s_{x_ia})\right] + \frac{|s_{z_i1}'|}{\sum^p_{k = 1}|s_{z_ik}'|}(\alpha)\right) + \ldots \\
&\quad\quad\quad\ldots + \left(\mu_fs_{y_ip} + s_{x_ip}\mu_g + \frac{1}{2}\left[\sum^p_{a = 1}(s_{x_ip}s_{y_ia} + s_{y_ip}s_{x_ia})\right] + \frac{|s_{z_ip}'|}{\sum^p_{k = 1}|s_{z_ik}'|}(\alpha)\right) \\
&= \mu_h + \frac{\sum^p_{k = 1}|s_{z_ik}'|}{\sum^p_{k = 1}|s_{z_ik}'|}(\alpha) + \sum^p_{j = 1}\left(\mu_fs_{y_ij} + s_{x_ij}\mu_g + \frac{1}{2}\left[\sum^p_{a = 1}(s_{x_ij}s_{y_ia} + s_{y_ij}s_{x_ia})\right]\right) \\
& = \mu_h + \alpha + \sum^p_{j = 1}\left(\mu_fs_{y_ij} + s_{x_ij}\mu_g + \frac{1}{2}\left[\sum^p_{a = 1}(s_{x_ij}s_{y_ia} + s_{y_ij}s_{x_ia})\right]\right)
\end{align*}

At this point we recall the definition given in Section \ref{subsec:math2} that $\mu_f\mu_g - \mu_h = \alpha$.
With a simple manipulation, we see that $\mu_h + \alpha = \mu_f\mu_g$. 
Thus,

\begin{align*}
    & = \mu_f\mu_g + \left(\mu_fs_{y_i1} + s_{x_i1}\mu_g + \frac{1}{2}\left[\sum^p_{a = 1}(s_{x_i1}s_{y_ia} + s_{y_i1}s_{x_ia})\right]\right) + \ldots \\
    &\quad\quad\quad\ldots + \left(\mu_fs_{y_ip} + s_{x_ip}\mu_g + \frac{1}{2}\left[\sum^p_{a = 1}(s_{x_ip}s_{y_ia} + s_{y_ip}s_{x_ia})\right]\right) \\
    &= \mu_f\mu_g + \sum^p_{j = 1}\mu_fs_{y_ij} + \sum^p_{j=1}s_{x_ij}\mu_g + \frac{1}{2}\sum^p_{j = 1}\sum^p_{a = 1}(s_{x_ij}s_{y_ia} + s_{y_ij}s_{x_ia}). \\
\end{align*}
We can expand this further to give us
\begin{align*}
    &= \mu_f\mu_g + \sum^p_{j = 1}\mu_fs_{y_ij} + \sum^p_{j=1}s_{x_ij}\mu_g + \frac{1}{2}(s_{x_i1}s_{y_i1} + s_{y_i1}s_{x_i1} + s_{x_i1}s_{y_i2} + s_{y_i1}s_{x_i2} + \ldots + s_{x_i1}s_{y_ip} \\
    &\quad\quad\quad + s_{y_i1}s_{x_ip} + s_{x_i2}s_{y_i1} + s_{y_i2}s_{x_i1} + s_{x_i2}s_{y_i2} + s_{y_i2}s_{x_i2} + \ldots + s_{x_i2}s_{y_ip} + s_{y_i2}s_{x_ip} \\
    &\quad\quad\quad + \ldots + \ldots \\ 
    &\quad\quad\quad+ s_{x_ip}s_{y_i1} + s_{y_ip}s_{x_i1} +s_{x_ip}s_{y_i2} + s_{y_ip}s_{x_i2} + \ldots + s_{x_ip}s_{y_ip} + s_{y_ip}s_{x_ip})\\
    &= \mu_f\mu_g + \sum^p_{j = 1}\mu_fs_{y_ij} + \sum^p_{j=1}s_{x_ij}\mu_g + \frac{1}{2}(2s_{x_i1}s_{y_i1} + 2s_{x_i1}s_{y_i2} + 2s_{x_i1}s_{y_i3} + \ldots + 2s_{x_i1}s_{y_ip}\\
    &\quad\quad\quad+ 2s_{x_i2}s_{y_i1} + 2s_{x_i2}s_{y_i2} + 2s_{x_i2}s_{y_i3} + \ldots + 2s_{x_i2}s_{y_ip} \\
    &\quad\quad\quad+ \ldots + \ldots \\
    &\quad\quad\quad+ 2s_{x_ip}s_{y_i1} + 2s_{x_ip}s_{y_i2} + 2s_{x_ip}s_{y_i3} + \ldots + 2s_{x_ip}s_{y_ip} \\
    &= (\mu_f + s_{x_i1} + s_{x_i2} + \ldots + s_{x_ip})(\mu_g + s_{y_i1} + s_{y_i2} + \ldots + s_{y_ip}).
\end{align*}

Since the original SHAP values have the local accuracy property, we know that
\[
(\mu_f + s_{x_i1} + s_{x_i2} + \ldots + s_{x_ip})(\mu_g + s_{y_i1} + s_{y_i2} + \ldots + s_{y_ip}) = \hat{x_i}\hat{y_i}
\]
Which in turn is equal to $\hat{z_i}$.
We see that $\hat{z_i} = \mu_h + \sum^p_{j = 1}s_{z_ij}$ and that the local accuracy property holds for mSHAP.

\section{The Simulation}
\label{methods:simulation}

\subsection{Simulation Process}

The basic flow for the simulation involved creating a data frame with all our desired combinations of $y_1$, $y_1$, $\theta_1$ and $\theta_2$ and then mapping through the following steps for each row:

\begin{enumerate}
    \item Using randomly distributed data as the covariates, create the response variables by evaluating $y_1$ and $y_2$ and then multiplying them together.
    \item Create two gradient boosted forests, one to predict $y_1$ and the other to predict $y_2$, based on the covariates.
    \item Multiply the model predictions together and run kernelSHAP to approximate explanations for the final model output.
    \item Use TreeSHAP to obtain exact explanations for the predictions of $y_1$ and $y_2$.
    \item Multiply the TreeSHAP values together, using the method described in Section \ref{sec:math} to calculate mSHAP values for each variable.
    \item Distribute $\alpha$ into the subsequent mSHAP values, in each of the four proposed ways.
    \item Compare the mSHAP values to the kernelSHAP values, using the scoring metrics described in Section \ref{subsec:sim1}.
    \item Record the resulting scores in a data frame.
\end{enumerate}

As previously mentioned, final scores were calculated by taking the average across all variables and all combinations of the inputs.
The code used to perform the simulation can be found in the github repo at https://github.com/srmatth/CAS, inside the mSHAP directory.

\subsection{Additional Simulations}
Since the initial simulation only used data with three explanatory variables, we have completed additional simulations with different numbers of variables.
The goal of this is to ascertain that the weighted by absolute value is the best method no matter the number of variables.

Our additional simulations used between 10 and 50 covariates across over 250 combinations of $y_1$, $y_2$, $\theta_1$, and $\theta_2$.
For these simulations all of our covariates were distributed uniformly between -1 and 1.
After performing the simulation, we saw that the absolute value method of weighting alpha is again the best (but just barely) based on overall score and in other metrics as well.

{\small
\begin{center}
\begin{tabular}{c|r r r r r r}
Method & Score & \makecell{Direction\\Score} & \makecell{Relative\\Value Score} &  Rank Score & \makecell{Pct Same\\Sign} & \makecell{Pct Same\\Rank}\\
\hline
&&&\\
\makecell{Weighted by\\Absolute Value} & 2.13 & 0.884 & 0.770 & 0.480 & 74.4\% & 24.9\% \\
&&&\\
\makecell{Uniformly\\Distributed} & 2.13 & 0.890 & 0.766 & 0.470 & 75.0\% & 23.7\% \\
&&&\\
\makecell{Weighted by\\Squared Value} & 2.12 & 0.880 & 0.768 & 0.475 & 73.9\% & 24.3\% \\
&&&\\
\makecell{Weighted by\\Raw Value} & 2.00 & 0.780 & 0.753 & 0.468 & 63.5\% & 23.2\% \\
\end{tabular}
\end{center}
}

\begin{figure}
    \centering
    \includegraphics[width=13cm]{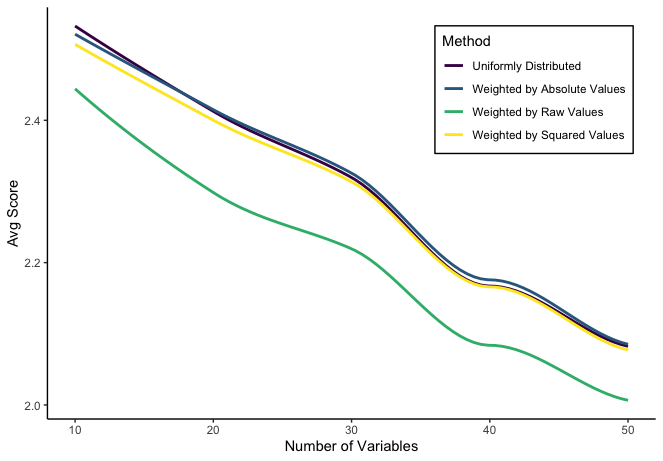}
    \caption{How the number of covariates impacts overall score, on average for mSHAP compared to kernelSHAP}
    \label{fig:n_var}
\end{figure}

\begin{figure}
    \centering
    \includegraphics[width=13cm]{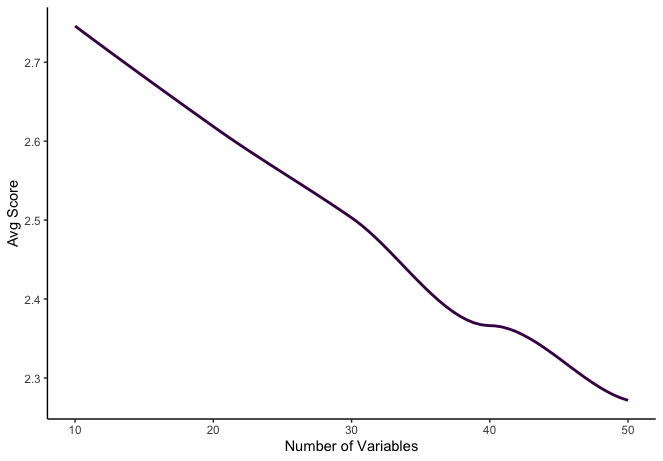}
    \caption{How the number of covariates impacts overall score, on average, for TreeSHAP compared to kernelSHAP}
    \label{fig:n_var_kernel}
\end{figure}

Due to these results, we are assured that the absolute weighting method of distributing $\alpha$ is the best based on our chosen metrics, across different numbers of covariates.
It can be seen in Figure \ref{fig:n_var} that the general score decreases as we add more variables.
However, this is consistent with what we see when we compare TreeSHAP (exact) to kernelSHAP (on singular models, not two-part models), as demonstrated in Figure \ref{fig:n_var_kernel}.

\section{The Data}
\label{methods:data}

The data used to create the model is a Property Damage data set, which is not available publicly but can be obtained through the Casualty Actuarial Society.

\section{The Model}
\label{methods:model}

Both the severity model and the frequency model were tuned in R using a h2o backend \citep{h2o_package_or_module}.
Tuning parameters are given in Table \ref{tab:tune} and model metrics are given in Table \ref{tab:metrics}.
Note that all model metrics were computed on the test (hold-out) subset of data.
These tuning results were then used to create the final model in python using scikit-learn \citep{scikit-learn}.
Note that scikit-learn was used to create the models because multinomial predictions do not have SHAP support in H2O as of the time of writing.


\begin{table}[h]
    \centering
    \begin{tabular}{c c c}
        \toprule
         Tuning Parameter & Severity Model & Frequency Model\\
         \toprule
         ntrees & 200 & 100 \\
         max\_depth & 30 & 20 \\
         mtries & 20 & 20 \\
         min\_split\_improvement & .0001 & .001 \\
         sample\_rate & 0.632 & 0.632 \\
         \bottomrule
    \end{tabular}
    \caption{Tuning parameters for the frequency and severity models}
    \label{tab:tune}
\end{table}

\begin{table}[h]
    \centering
    \begin{tabular}{c c c c}
        \toprule
         Model & MAE & MSE & Logloss \\
         \toprule
         Severity Model & 2,832 & 16,359,170 & NA\\
         Frequency Model & NA & 0.074 & 0.427\\
         \midrule
         Two-Part Model & 683 & 830,351 & NA\\
         \bottomrule
    \end{tabular}
    \caption{Model metrics for all models}
    \label{tab:metrics}
\end{table}

\section{Code Availability}
\label{methods:code}

The code used to tune the model (as well as additional code focused on working with the CAS datasets) can be found at this github link: https://github.com/srmatth/CAS.

mSHAP has been developed into an R package as well.
The R package can be downloaded from CRAN, with the R code
\begin{lstlisting}
install.packages("mshap")
\end{lstlisting}
or get the development version from https://github.com/srmatth/mshap by running
\begin{lstlisting}
devtools::install_github("srmatth/mshap")
\end{lstlisting}
in R.

The mSHAP package repository (https://github.com/srmatth/mshap) also contains all the code and data used to generate the plots in this paper, as well as the code used to run the various simulations mentioned.
It can be found in the inst/paper directory under the main directory of the package.
Be aware that installing the package by following the steps above will not download the code used in this paper, it must be obtained from the github repository.

\vspace{12pt}
\noindent\textbf{Acknowledgments}
\begin{itemize}
    \item Brigham Young University Department of Statistics Computing Cluster
    \item Brian Fanin and the Casualty Actuarial Society for providing the data
    \item Isabelle Matthews for proof-reading
\end{itemize}

\noindent Funding: This paper was funded by an individual grant from the Casualty Actuarial Society.

 \bibliographystyle{elsarticle-harv} 
 \bibliography{mshap}





\end{document}